\documentclass[preprint,10pt]{elsarticle}
\usepackage[T1]{fontenc}
\usepackage{lmodern}
\usepackage{graphicx}
\usepackage{booktabs}
\usepackage{amsmath}
\usepackage{amssymb}
\usepackage[section]{placeins} 
\usepackage{hyperref} 
\usepackage{easyReview}
\newcommand{\Vis}[1]{\mathrm{Vis}_P(#1)}
\newcommand{\Covf}[1]{\mathrm{Cov}_P(#1)}
\newcommand{\Area}[1]{\mathrm{Area}(#1)}
\newcommand{\EOS}{\mathrm{EOS}}
\newcommand{\OPT}{\mathrm{OPT}}
\newcommand{\SetPredictor}{\textup{\textsc{SetPredictor}}}
\newcommand{\cfeas}{0.95} 

\newcommand{\figincl}[4][\linewidth]{%
  \begin{figure}[t]
    \centering
    \includegraphics[width=#1]{#2}%
    \caption{#3}
    \label{#4}
  \end{figure}%
}

\begin{document}

\begin{frontmatter}

\title{Learning to Place Guards by Reinforcement:\\
A Geo-Free Neural Policy for the Vertex-Guard\\
Art Gallery Problem}

\author[osijek]{Domagoj Ševerdija\corref{cor1}}
\ead{dseverdi@mathos.hr}
\author[osijek]{Jurica Maltar}
\ead{jmaltar@mathos.hr}
\author[]{Nathan Chappel}
\ead{nathan.s.chappel@gmail.com}
\author[osijek]{Domagoj Matijević}
\ead{domagoj@mathos.hr}

\cortext[cor1]{Corresponding author}
\address[osijek]{School of Applied Mathematics and Informatics,
  University J.~J.~Strossmayer of Osijek, Croatia}

\begin{abstract}
Neural combinatorial optimization (NCO) has shown that policies trained by reinforcement can construct strong solutions to NP-hard problems directly from raw instances. What such a policy actually \emph{learns}, as opposed to what its decoder \emph{expresses}, remains much less clear. We study this distinction on the vertex-guard Art Gallery Problem, the NP-hard task of choosing polygon vertices from which to observe an entire region. A pointer-network policy is trained from a coverage-aware reward over its own rollouts under the constraint we call \emph{geo-free} inference: at test time it sees only vertex coordinates, with no visibility computation and no geometric oracle. The policy places guards economically but leaves a tail of under-covered polygons that widens far beyond the training range. To locate the cause, we freeze the trained encoder and read its embeddings with a small single-shot classifier, still geo-free at inference. The classifier closes most of the feasibility gap, in and out of distribution and at up to roughly five times the training range, cutting under-covered polygons by about an order of magnitude at an explicitly reported cost in guard count. We read this as evidence that the reinforcement-trained representation already encodes the geometry required for feasibility, and that residual failures reflect decoder calibration rather than missing knowledge. Probing a frozen encoder thus offers a practical way to ask what a neural combinatorial solver has internalized.
\end{abstract}


\end{frontmatter}

\newpage

\section{Introduction}
\label{sec:intro}

The Art Gallery Problem (AGP) asks for the minimum number of guards needed to observe every point of a simple polygon~\cite{LeeLin1986,ORourke1987AGT}, with applications in surveillance, sensor placement, and robotics~\cite{1545170,DBLP:journals/evi/BighamDK22}. We focus on the \emph{vertex-guard} variant, in which guards must be chosen from the polygon's vertices; the problem remains NP-hard and is a discrete instance of geometric set cover with non-uniform, polygon-dependent visibility regions. The question this paper investigates is whether a neural policy can learn to place such guards from coordinates alone, trained from a coverage-aware reward signal over its own rollouts, with no visibility matrix and no geometric oracle at the moment it chooses guards. We are not asking whether a learned solver can match a classical greedy heuristic on guard count --- it cannot, and we do not claim it does. We are asking whether the underlying combinatorial-geometric task is learnable when the model is denied any explicit geometric input at test time.

The reinforcement method we evaluate is an LSTM pointer-network policy trained with Preference Optimization (PO)~\cite{Pan2025PO} on Bradley--Terry preference pairs~\cite{BradleyTerry1952} drawn from the policy's own rollouts. The policy emits a vertex-index sequence with an $\EOS$ token; the reward couples coverage and guard usage. PO replaces REINFORCE advantages~\cite{Williams1992,Bello2016} with binary preferences between pairs of policy rollouts, preserving a usable gradient signal even when the reward saturates under the coverage constraint~\cite{Pan2025PO}. We refer to inference under this policy as \emph{geo-free} --- the network reads only vertex coordinates at the moment it places guards, never querying a visibility oracle (defined precisely in Section~\ref{sec:problem-geofree}).

The policy works, to a degree. On a held-out in-distribution split its guard sets are competitive in cardinality with classical baselines, but the per-polygon coverage distribution has a tail: a non-trivial fraction of polygons end up below the feasibility line one would want a deployed solver to clear, and the failure rate grows on out-of-distribution polygons whose vertex counts exceed the training range. The reinforcement method, by itself, is not feasibility-guaranteed.

To check whether this residual reflects a true limit of the reinforcement-learned representation or only a calibration problem in the decoder, we train a small per-vertex classifier conditioned on the policy's frozen encoder embeddings --- together with the vertex coordinates and a binary indicator of the policy's own guard set, all geo-free --- and predict a vertex-inclusion probability, controlled by a single scalar threshold. The classifier has no access to visibility information at inference, and it compresses the feasibility tail substantially both in and out of distribution. A no-encoder ablation that zeroes the embeddings while keeping the coordinates and seed indicator (Section~\ref{sec:res-headline-ablation}) isolates the encoder's contribution from these other inputs, and we read the gap as evidence that the reinforcement-trained encoder already carries enough geometric structure to support a feasibility decision --- so the policy's coverage tail reflects what its decoder expressed, not a limit of the representation. The cost of recovering feasibility is a controlled rise in guard count, which we report explicitly.

\paragraph{Contributions} This paper makes four contributions.
\begin{itemize}
  \item[\textbf{C1}] \textbf{A representation--decoder decomposition for neural combinatorial optimization.} We adapt the probing methodology of representation analysis~\cite{AlainBengio2017Probes,HewittLiang2019Probes,Belinkov2022Probing} to a reinforcement-trained combinatorial solver: freezing the policy's encoder and training a single-shot per-vertex probe over its embeddings (with the vertex coordinates and the policy's seed) isolates what the learner \emph{represents} from what its decoder \emph{expresses}, with a $2\times2$ encoder--seed ablation and a zero-capacity linear probe as controls. To our knowledge this is the first such probing study of an RL-trained policy on a geometric covering problem.
  \item[\textbf{C2}] \textbf{Out-of-distribution generalization at scale.} On a large held-out set spanning the training size range and extending to roughly five times beyond it, the probe raises the average fraction meeting the $\cfeas$ feasibility gate across four probe seeds from $85\%$ (policy seed) to $99\%$; the gain holds on the genuinely larger-than-training polygons alone ($85\% \to 98.7\%$, Section~\ref{sec:res-ood}).
  \item[\textbf{C3}] \textbf{Geo-free inference as a protocol.} Restricting test-time inputs to vertex coordinates and learned features derived from them --- never a visibility oracle --- isolates what a learner internalizes from what such an oracle could compute for it, making ``no explicit geometric knowledge'' a measurable property rather than a rhetorical one (Section~\ref{sec:problem-geofree}).
  
  \item[\textbf{C4}] \textbf{A structural finding on post-processing.} Learned iterative editors fail to improve the policy's seed without dropping coverage --- due to error accumulation, stop-time miscalibration, and train/inference drift --- whereas the single-shot probe avoids all three by design and is empirically a fixed point: re-running it on its own output changes nothing (Section~\ref{sec:res-iteration}). That contrast validates the single-shot design as a clean reading of the encoder.
\end{itemize}
We do not claim to beat classical greedy or local-search heuristics on guard count; we claim, and document, that vertex-guard placement is learnable to a measurable degree by a reinforcement method that never reads an explicit visibility object at inference.

The remainder of the paper is organized as follows. Section~\ref{sec:problem} establishes notations and definitions. In Section~\ref{sec:related} related work is reviewed. Section~\ref{sec:method} describes the reinforcement method and the representation probe. Finally, Section~\ref{sec:experiments} describes the experimental setup, Section~\ref{sec:results} reports the results, followed by discussion in Section~\ref{sec:discussion}, limitations in Section~\ref{sec:limitations}, and the conclusion in Section~\ref{sec:conclusion}.

\section{Problem and Notation}
\label{sec:problem}

\subsection{Polygons and visibility}
\label{sec:problem-polygon}

Let $P \subset \mathbb{R}^2$ denote a simple polygonal region (the closed set consisting of its interior and boundary), with $n$ vertices at coordinates $x_1, \dots, x_n \in \mathbb{R}^2$ and vertex index set $V(P) = \{1, \dots, n\}$. For two points $p,q\in P$  we say that $p$ \emph{sees} $q$ when the closed segment between them stays inside the polygon, $\overline{pq} \subseteq P$. Guards are placed at vertices: for a vertex-guard set $S \subseteq V(P)$, the visibility region and area-normalized coverage are
\begin{equation}
\Vis{S} \,=\, \{\, q \in P : \exists\, i \in S,\ x_i \text{ sees } q \,\},
\qquad
\Covf{S} \,=\, \frac{\Area{\Vis{S}}}{\Area{P}} \in [0,1].
\label{eq:vis-cov}
\end{equation}

\subsection{The vertex-guard variant}
\label{sec:problem-vertex}

The \emph{vertex-guard Art Gallery Problem (AGPVG)} asks for the smallest vertex-guard set that covers the whole polygon. We define the \emph{vertex-guard optimum} directly:
\begin{equation}
\OPT(P) \,=\, \min\bigl\{|S| : S \subseteq V(P),\ \Covf{S} = 1\bigr\}.
\label{eq:opt}
\end{equation}
Restricting guards to $V(P)$ reduces placement to a finite discrete choice over $n$ vertices --- the natural action space for a pointer-network policy~\cite{Vinyals2015,Bello2016,Kool2019Attention,Kwon2020POMO} --- while retaining the full NP-hardness and non-uniform visibility structure of the problem.

Note that equation~\eqref{eq:opt} is a \emph{constrained} single-objective problem --- minimize the guard count subject to the hard constraint $\Covf{S} = 1$ --- and is a discrete set-cover instance, where coverage is the submodular union of per-guard visibility regions. This is what separates it from the routing problems on which neural combinatorial solvers are usually studied, where any permutation is feasible and only a single scalar cost is minimized. A learner denied a visibility oracle at inference (the geo-free constraint, Section~\ref{sec:problem-geofree}) cannot certify the coverage constraint while placing guards; it must instead trade coverage against guard count, so the single constrained objective becomes a two-axis operating curve. 

\subsection{Geo-free inference}
\label{sec:problem-geofree}

We will say that an inference procedure is \emph{geo-free} if its inputs at test time are limited to the polygon's vertex coordinates and learned features derived from them --- no polygon visibility matrix, no CGAL-computed visibility regions, no per-vertex area or visibility-fraction feature. Geo-free inference does not preclude using exact visibility during evaluation (to report coverage after the fact) or during training (to compute rewards, gate trajectories, or build supervised targets); it restricts what the model reads at the moment it places guards. A solver allowed to query a visibility oracle at every decision step --- as classical local search and active-search decoding both do --- can in principle simulate greedy or local search, and a measurement of what such a solver has learned would be conflated with a measurement of what its oracle can compute. The geo-free constraint isolates the first question from the second, which is what makes ``no explicit geometric knowledge'' a measurable property.

\section{Related Work}
\label{sec:related}

This section places the present work in four lines of research: algorithms for the Art Gallery Problem (classical and learned), neural combinatorial optimization with reinforcement learning, supervised post-processing of learned solvers, and probing of learned representations.

\paragraph{The Art Gallery Problem} Lee and Lin~\cite{LeeLin1986} established NP-hardness for several variants of the AGP; O'Rourke~\cite{ORourke1987AGT} surveys the foundational geometric and combinatorial results. Work on the problem has traditionally followed three lines. Exact algorithms solve smaller instances to optimality, most effectively via integer programming: Couto et al.~\cite{Couto2011ExactAGP} compute exact vertex-guard optima this way, and we treat their published instance library~\cite{art-gallery-instances-page}, with its exact optima, as the source of polygons in our experiments. Approximation algorithms provide provable guarantees --- Ghosh~\cite{Ghosh2010ApproxAGP} gives a logarithmic-factor method for guarding simple polygons. Heuristics without theoretical guarantees, of which the classical greedy rule~\cite{GonzalezBanos2001} that places guards to maximize marginal coverage is the most common, are reliable on small to mid-sized polygons. Because area coverage is a monotone submodular function of the guard set, this marginal-coverage rule is exactly the greedy submodular-maximization heuristic and inherits its diminishing-returns approximation guarantee~\cite{NemhauserWolseyFisher1978}; the guarantee is available only to a solver that can evaluate marginal coverage while it selects, which a geo-free learner cannot (Section~\ref{sec:problem-geofree}), so we forgo such guarantees by construction --- the price of measuring what the representation has learned rather than what an oracle computes. The same exact/approximation/heuristic split recurs across the problem's many variants --- for instance, guarding $1.5$-dimensional terrains admits constant-factor approximations~\cite{journals/siamcomp/Ben-MosheKM07,EMMS08}. Local-search refinement of a candidate guard set by swapping or removing vertices is the standard high-quality post-processing step. We use classical greedy purely as a non-learned evaluation reference, not in competition with the reinforcement method on guard count.

Reinforcement learning has been applied to the AGP and to closely related coverage problems before, but in settings quite different from ours. Liao et al.~\cite{Liao2025AGPRL} discretize the polygon into a grid and train a per-instance tabular Q-learning agent that trades guard count against covered area; Kaba et al.~\cite{Kaba2016ViewPlanning} cast the 3D view-planning problem --- a sensor-coverage set-cover --- as an MDP solved with value-based RL (SARSA, Q-learning, temporal-difference) guided by a geometry-aware score, outperforming the greedy baseline. Both methods consult coverage or visibility while placing guards and learn a fresh policy for each instance. We differ on three axes: the action space is the polygon's own vertices rather than grid cells or candidate viewpoints; a single amortized policy generalizes to unseen polygons without retraining; and inference is geo-free, with no visibility computation at the moment guards are placed.

\paragraph{Reinforcement learning for combinatorial optimization} Pointer networks~\cite{Vinyals2015} introduced the architecture of attending over input tokens to produce variable-length output index sequences, and were extended to combinatorial optimization with reinforcement learning by Bello et al.~\cite{Bello2016} using REINFORCE~\cite{Williams1992} with a learned baseline; Deudon et al.~\cite{Deudon2018TSP} similarly train TSP construction heuristics by policy gradient. Subsequent work introduced attention-only architectures for routing problems~\cite{Kool2019Attention}, policy-optimization variants such as POMO~\cite{Kwon2020POMO} that exploit problem symmetries, and Transformer pointer policies such as Pointerformer~\cite{Jin2023Pointerformer} aimed at generalizing to large instances --- a concern we share, since our evaluation stresses out-of-distribution polygon sizes. We adopt the more recent preference-optimization (PO) recipe of Pan et al.~\cite{Pan2025PO}, which trains the policy with a Bradley--Terry~\cite{BradleyTerry1952} ranking loss over pairs of solutions sampled from the policy itself, replacing the variance-heavy REINFORCE objective. We treat PO as a reinforcement-style procedure: the loss is computed from policy rollouts on the environment (the polygon), the gradient direction is determined by the reward (coverage versus guard usage), and no supervised labels for the action sequence are used. A broader methodological survey is provided by Bengio et al.~\cite{survey:BENGIO2021405}.

A recurring feature of this literature is worth making explicit, because AGP departs from it. Routing problems such as the TSP carry a single scalar objective and free feasibility --- any permutation is a valid tour --- and even constrained variants such as capacitated routing usually enforce feasibility by masking infeasible actions at decode time, which presumes an oracle for the constraint. Vertex-guard AGP has neither property: it is a constrained set-cover problem (Section~\ref{sec:problem-vertex}), and under geo-free inference the coverage constraint cannot be masked in, so it can only be learned and traded off against guard count --- the operating curve our threshold sweep traces. The closest learned analogues are RL methods that approximate Pareto frontiers for bi-objective routing~\cite{Mehta2022PANet,Li2019DRLMOA}; but those balance two \emph{soft} costs, whereas one of our two axes is a hard feasibility constraint, making PA-Net's Lagrangian relaxation of a constraint~\cite{Mehta2022PANet} a closer match than a symmetric Pareto front. On the optimization side, Caramanis et al.~\cite{Caramanis2023Landscape} prove that policy-gradient solution-samplers enjoy benign landscapes for a broad class of problems (Max-/Min-Cut, Max-$k$-CSP, bipartite matching, TSP), but their guarantee assumes a single cost that is \emph{bilinear} in instance and solution features and imposes no hard feasibility constraint. AGP's guard-count term is linear and would fit, but its coverage term is the submodular union of visibility regions and its full-coverage requirement is a hard, geo-free-uncertifiable constraint --- placing AGP outside their analyzed class and helping explain why a naive policy gradient struggles here (Section~\ref{sec:method-pointer}).

\paragraph{Supervised post-processing of learned solvers} An orthogonal line of work post-processes a base solver's output with a second model trained by supervised or imitation learning. For combinatorial problems with hard feasibility constraints (such as covering), iterative supervised editors must contend with compounding errors and a train--inference distribution gap; DAgger~\cite{RossGordonBagnell2011DAgger} and its variants partially address the latter. We use a single-shot supervised post-processor (the \SetPredictor{}) not as a competing solver but as a \emph{probe} of the RL-trained encoder's representation. The framing is methodological: if a small classifier reading frozen RL embeddings can recover feasibility, the RL encoder must already carry the relevant geometric information. Directly cloning search solutions with a recurrent decoder, by contrast, fails on this problem for a structural reason --- the teacher-forcing cascade we return to in Section~\ref{sec:res-iteration} --- which is why we read the frozen encoder with a single-shot, recurrence-free probe rather than a sequential supervised decoder.

\paragraph{Probing learned representations} The methodology of freezing a trained network and reading its internal representations with a small supervised classifier originates in the analysis of vision and language models: linear classifier probes diagnose what intermediate layers encode~\cite{AlainBengio2017Probes}, control tasks calibrate how much of a probe's accuracy reflects the representation rather than the probe's own capacity~\cite{HewittLiang2019Probes}, and Belinkov~\cite{Belinkov2022Probing} surveys the family. We adapt this methodology to neural combinatorial optimization, where it has seen little use: the probed model is a \emph{reinforcement-trained} policy rather than a self-supervised encoder, the probed property is a geometric feasibility signal (guard membership) rather than linguistic structure, and the geo-free constraint fixes what the probe may read at inference. Our no-encoder ablation (Section~\ref{sec:res-headline-ablation}) plays the role of the control task: it keeps the probe's capacity and training recipe constant while removing the representation, so the performance gap is attributable to the encoder.

\section{Method}
\label{sec:method}

We describe the reinforcement method (Section~\ref{sec:method-pointer}) and the single-shot representation probe (Section~\ref{sec:method-arch}). The empirical diagnostic that most motivates the single-shot design, i.e., why iterative editors fail to improve the seed, is deferred to the results (Section~\ref{sec:res-iteration}).

\subsection{The reinforcement method: PO/BT pointer policy}
\label{sec:method-pointer}

\paragraph{Architecture} The policy is a compact pointer network with a single-layer, unidirectional LSTM~\cite{Hochreiter1997LSTM} encoder and an attention-based decoder. A polygon $P$ with vertex coordinates $(x_1, \dots, x_n)$ is consumed coordinate-wise (no visibility input). Running the LSTM over the vertex sequence yields per-vertex hidden states that serve as the embeddings $(h_1, \dots, h_n)$, $h_i \in \mathbb{R}^{d}$, where $d$ is the encoder embedding dimension (its value, and those of all dimensions introduced below, is set in Section~\ref{sec:exp-training}). The decoder attends over the encoder at each step, emits a vertex index or an \emph{end-of-sequence} -  $\EOS$ token, and stops at $\EOS$ or when the maximum length is reached. The decoder outputs a sequence $\pi = (\pi_1, \dots, \pi_{|\pi|})$ while the resulting guard set is $S(\pi) = \{\pi_t : \pi_t \ne \EOS\}$. All metrics, i.e., coverage, $|S|/n$, $|S|/\OPT$, are computed from $S(\pi)$. We denote the joint encoder--decoder distribution by $p_\theta(\pi \mid P)$, where $\theta$ collects all policy parameters.

The encoder has no training signal of its own: its weights change only because the reward gradient back-propagates from the decoder's rollouts. If the encoder already holds the geometric information needed for feasibility, a small classifier reading its frozen embeddings should be able to recover the guard set the decoder failed to express. This is why we probe the encoder inside the trained policy by freezing it and attaching a small classifier. By contrast, a standalone model, i.e., an encoder + a probe trained from scratch, would reflect its own learning, not what the reinforcement policy internalized. The policy's greedy decode additionally supplies the seed guard set that the representation probe (Section~\ref{sec:method-arch}) later refines.

\paragraph{Why an autoregressive pointer} Vertex-guard placement is a set-cover problem: whether a vertex should be a guard depends on what the guards already chosen leave uncovered. An autoregressive pointer captures this dependency naturally by conditioning each pick on the partial solution (Eq.~\eqref{eq:lognorm}), and it decides cardinality through the $\EOS$ token rather than a fixed threshold. A model scoring vertices independently would miss this structure --- which is also why the probe (Section~\ref{sec:method-arch}) is not a standalone solver: unable to be autoregressive, it takes the policy's greedy seed as an input instead.

The reason the decoder is necessary once the probe works, i.e., why we cannot discard it, is that the encoder's representation was shaped entirely by the reward gradient flowing through the decoder, and the probe reads the decoder's seed as a feature. Removing the decoder removes both the representation and the probe's input. What we compare is encoder-and-decoder against encoder-and-probe, not the decoder against no-decoder. Reading the frozen encoder with a probe also recovers feasibility on polygons far larger than those seen in training (Section~\ref{sec:res-ood}), which autoregressive pointer policies alone cannot do~\cite{Vinyals2015,Jin2023Pointerformer}.

\paragraph{Reward and rollouts} For each polygon $P$ we sample $R$ rollouts from the policy and score each rollout's guard set $S(\pi)$ by a scalar utility that gates coverage at $\tau$ and then prefers smaller guard sets,
\begin{equation}
u(\pi \mid P) \,=\, \min\!\big(\Covf{S(\pi)},\, \tau\big) \;-\; \lambda\,\frac{|S(\pi)|}{|V(P)|} \;-\; \rho\,\max\!\big(0,\, \tau - \Covf{S(\pi)}\big),
\label{eq:reward}
\end{equation}
where $\tau = 0.99$ is the coverage gate, $\lambda = 1.0$ weights guard usage, and $\rho = 3.0$ penalizes coverage below the gate (Section~\ref{sec:exp-training}). Capping the coverage term at $\tau$ removes any incentive to over-cover, so once a rollout is feasible the utility is driven purely by cardinality. Coverage during training is computed with the discretized-visibility sampler described next ($M = 500$ interior points per polygon), a cheap approximation of the exact coverage area.

\paragraph{Discretized-visibility coverage} Computing the exact coverage of a guard set --- the area of the union of its vertices' visibility polygons --- is too costly to repeat for the many rollouts and local-search candidates scored per polygon, so during training we estimate it by Monte-Carlo sampling. For each polygon we draw $M = 500$ points uniformly from its interior (rejection sampling within the bounding box, under a fixed per-polygon seed) and, for each vertex $i$, compute its \emph{exact} visibility polygon $\Vis{\{i\}}$ once with CGAL triangular-expansion visibility~\cite{cgal:eb-25b}. Recording which samples fall inside each visibility polygon yields a boolean visibility matrix $B \in \{0,1\}^{|V(P)| \times M}$ with $B_{is} = 1$ iff sample $s$ is visible from vertex $i$, and the coverage of a guard set $S$ is estimated by the covered-sample fraction
\begin{equation}
\overline{\Covf{S}} \,=\, \frac{1}{M}\,\Big|\big\{\, s : \exists\, i \in S,\ B_{is} = 1 \,\big\}\Big|,
\label{eq:discrete visibility}
\end{equation}
a uniform Monte-Carlo estimate of the area ratio $\Covf{S}$ of Eq.~\eqref{eq:vis-cov}, evaluated as a bitwise OR over the rows of $B$. Because $B$ depends only on the polygon, it is computed once and cached, so every subsequent coverage query during a rollout or a local-search edit costs an $O(|S|\,M)$ bitwise operation rather than a polygon-union computation. The per-vertex visibility polygons are exact; the only approximation is replacing the area integral by a sample average, whose standard error decays as $O(1/\sqrt{M})$. Exact coverage --- the area of the union, integrated rather than sampled --- is reserved for evaluation (Section~\ref{sec:exp-eval}).

\paragraph{Preference-optimization loss} We score a rollout by its \emph{length-normalized} log-likelihood --- the mean per-token log-probability under the factorization $$p_\theta(\pi \mid P) = \prod_t p_\theta(\pi_t \mid \pi_{<t}, P)$$,
\begin{equation}
\bar{\ell}_\theta(\pi \mid P) \,=\, \frac{1}{|\pi|} \sum_{t=1}^{|\pi|} \log p_\theta(\pi_t \mid \pi_{<t}, P),
\label{eq:lognorm}
\end{equation}
which keeps the preference from being biased by guard-set size, since AGP solutions have variable length (cf.\ Pan et al.~\cite{Pan2025PO}). For each pair $(\pi^+, \pi^-)$ of rollouts with $u(\pi^+ \mid P) > u(\pi^- \mid P)$ above a coverage gate, we apply the Bradley--Terry log-likelihood
\begin{equation}
\mathcal{L}_{\mathrm{BT}}(\theta) \,=\, -\mathbb{E}_{(P, \pi^+, \pi^-)} \left[\, \log \sigma\!\big( \alpha\, [\, \bar{\ell}_\theta(\pi^+ \mid P) - \bar{\ell}_\theta(\pi^- \mid P)\, ] \big) \,\right],
\label{eq:bt-loss}
\end{equation}
where $\alpha > 0$ is a temperature scaling the preference margin ($\alpha = 0.05$; Section~\ref{sec:exp-training}). The gradient direction is driven purely by which of the two rollouts achieved a higher reward; no supervised target sequence is used.

The choice of PO over REINFORCE is deliberate. Once a rollout meets the coverage constraint, the reward of Eq.~\eqref{eq:reward} flattens and the policy-gradient signal on guard count vanishes --- a saturation effect specific to hard coverage constraints, not covered by benign-landscape guarantees for unconstrained problems~\cite{Caramanis2023Landscape}. We observe this failure directly: a value-critic REINFORCE policy over-guards by roughly $3.5\times$ the optimum and a greedy-baseline variant by about $7\times$ (Table~\ref{tab:reinforce}). Bradley--Terry preferences between rollout pairs remain informative even when both rollouts saturate, since the signal depends on which rollout is relatively better, not on the absolute reward level. PO/BT reaches comparable coverage at $1.3\times$ the optimum. At the end of training the encoder is frozen and treated as a fixed feature extractor for the rest of the paper; Figure~\ref{fig:po_training} shows the late-phase policy holding a stable operating point --- good coverage at low guard cost --- rather than collapsing onto a single axis.

\begin{table}[t]
  \centering
  \small
\begin{tabular}{lrr}
  \toprule
  Training objective & Mean cov & Mean $|S|/\OPT$ \\
  \midrule
  REINFORCE, learned-critic baseline & 0.904 & 3.49 \\
  REINFORCE, greedy rollout baseline & 1.000 & 7.00 \\
  PO / Bradley--Terry (ours) & \textbf{0.978} & \textbf{1.30} \\
  \bottomrule
\end{tabular}

  \caption{Training-objective comparison on the combined \texttt{dev}$+$\texttt{test} splits (greedy decode). Coverage is geometric polygon coverage; $|S|/\OPT$ is the mean approximation ratio. The REINFORCE rows are $30$-epoch runs evaluated under an earlier version of the evaluation protocol; they are reported as evidence for the choice of training objective, not as tuned, protocol-matched competitors. The PO/BT row is the policy used throughout the paper.}
  \label{tab:reinforce}
\end{table}

\figincl[0.75\linewidth]{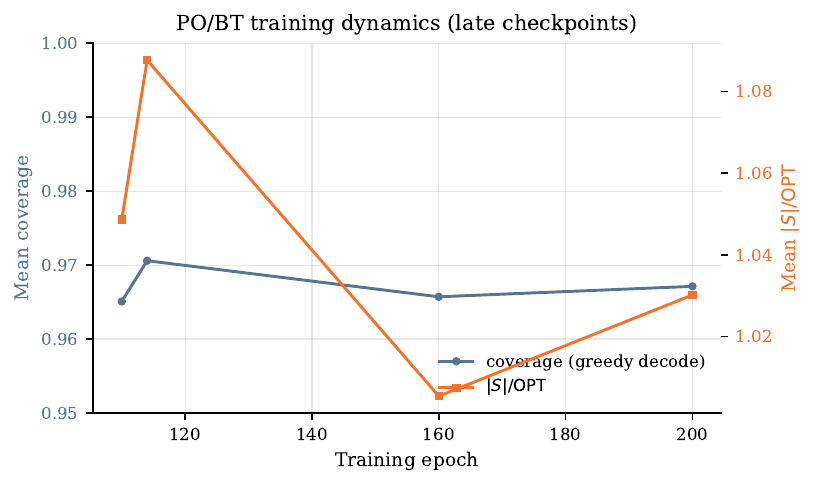}{PO/BT training dynamics over the four late checkpoints (epochs $110$, $114$, $160$, $200$), evaluated on a fixed held-out sample of $100$ polygons; per-epoch logging was not preserved, so early training is not shown (Section~\ref{sec:limitations}). Mean per-polygon coverage (left axis, zoomed to $[0.95, 1.0]$) and $|S|/\OPT$ (right axis) of the greedy-decoded policy stay within a narrow band with no sign of collapse; the movement is non-monotone rather than a clean trend, which is all four checkpoints can establish.}{fig:po_training}

\subsection{Probing the representation: a single-shot per-vertex classifier}
\label{sec:method-arch}

The policy learns to decrease the cardinality but leaves a feasibility tail (Section~\ref{sec:res-headline}). We first verified that this tail is \emph{reachable}: running a local-search editor (add/remove/swap) on the policy's own greedy seed restores feasibility on every sub-feasibility in-distribution polygon while \emph{reducing} the guard count (the \emph{LS on policy seed} row of Table~\ref{tab:headline}). We treat this as a sanity check that settles the diagnosis --- the missing feasibility is a handful of local edits away from the seed, so it reflects decoder calibration rather than absent geometry --- and that, as a by-product, yields a concrete target set $S^\star_{\mathrm{LS}}$. The open question is then whether that refinement can be \emph{learned} from the frozen encoder, geo-free. A learned \emph{iterative} editor over the seed cannot: it does not close the tail in any coverage-preserving regime (Section~\ref{sec:res-iteration}), because error accumulation, stop-time miscalibration, and train/inference drift make a rollout-based post-processor unreliable here, contrary to fine-tuning arguments of the policy in \cite{Pan2025PO}. We therefore probe what the policy has learned with a \emph{single-shot, rollout-free} classifier that recovers the feasibility decision the decoder failed to make. Its inputs are geo-free --- the frozen encoder embedding, the vertex coordinates, and the policy's seed indicator --- with no visibility object at inference. Having no rollout and no stop decision, a single pass has nothing to drift and nothing to miscalibrate. We call it the \SetPredictor{} and present it primarily as a representation diagnostic; the threshold sweep in Section~\ref{sec:results} makes the diagnostic concrete, but the operating-curve framing is a side-effect of the measurement, not a deployment recipe.

For polygon $P$ with vertex coordinates $x_1, \dots, x_n$, the input feature for vertex $i$ is
\begin{equation}
z_i \,=\, [\, h_i \,;\, x_i \,;\, \mathbf{1}\{i \in S(\pi_{\mathrm{seed}})\} \,] \,\in\, \mathbb{R}^{d+3},
\label{eq:setpred-feat}
\end{equation}
that is, the \emph{frozen} $d$-dimensional encoder embedding $h_i$ from the PO/BT policy, the 2D coordinates $x_i$, and a single binary indicator for whether vertex $i$ is selected by the policy's greedy-decoded seed $\pi_{\mathrm{seed}}$. 

The architecture is a small Transformer encoder (parameters $\phi$) over the $n$ vertex tokens with a sigmoid output head per token (sizes given in Section~\ref{sec:exp-training}). The features $z_i$ are linearly projected to width $d$ and passed through $L$ pre-norm self-attention blocks --- each with $H$-head self-attention at width $d$, a GELU feed-forward sublayer of expansion factor $f$, and residual connections --- followed by a final layer normalization; we write $v_i$ for the resulting per-vertex hidden state. This $v_i$ is the \SetPredictor{}'s \emph{own} encoding of vertex $i$, and is distinct from the frozen policy embedding $h_i$, which enters only as part of the input feature $z_i$ in Eq.~\eqref{eq:setpred-feat}. From the $(v_1, \dots, v_n)$ we build two context vectors by mean-pooling: a \emph{seed} context $c_{\mathrm{seed}} = \tfrac{1}{|S(\pi_{\mathrm{seed}})|}\sum_{i \in S(\pi_{\mathrm{seed}})} v_i$, averaged over the vertices the policy placed in its seed, and a \emph{global} context $c_{\mathrm{glob}} = \tfrac{1}{n}\sum_{i=1}^{n} v_i$, averaged over all vertices of the polygon. Both are single vectors shared across vertices. Each per-vertex prediction concatenates the vertex's own $v_i$ with these two context vectors and passes the result through a 2-layer GELU MLP head ($3d \to d \to 1$):
\begin{equation}
\hat{p}_i \,=\, \sigma\!\big( \mathrm{MLP}([\, v_i \,\|\, c_{\mathrm{seed}} \,\|\, c_{\mathrm{glob}} \,]) \big),\qquad i = 1, \dots, n.
\label{eq:setpred-arch}
\end{equation}
The output $\hat{p}_i \in [0,1]$ is interpreted as the probability that vertex $i$ should be in the final guard set. Thresholding at $t \in (0,1)$ yields the guard set
\begin{equation}
S(t) \,=\, \{\, i \in \{1, \dots, n\} : \hat{p}_i \geq t \,\}.
\label{eq:setpred-decode}
\end{equation}

\paragraph{Objective} The probe is trained by supervised learning to predict membership in an LS-improved target set $S^\star_{\mathrm{LS}}(P)$ --- the policy's seed refined by local search, constructed as described in Section~\ref{sec:exp-training} --- so it is not itself a reinforcement learner. We supervise on this refinement rather than on the exact optimum for two reasons: guard membership at the optimum is non-unique --- the constraint in Eq.~\eqref{eq:opt} fixes cardinality, not which vertices are chosen (Section~\ref{sec:res-iteration}) --- so an optimal set is an ill-posed per-vertex label, and the optimum is in any case unavailable at scale (Section~\ref{sec:exp-eval}); the LS endpoint, by contrast, is a well-defined, feasibility-restoring refinement of the policy's \emph{own} seed. With $y_i = \mathbf{1}\{i \in S^\star_{\mathrm{LS}}(P)\}$, a positive-class weight $w_{+}$ that balances the empirical non-guard-to-guard ratio, and $\mathcal{B}$ a mini-batch of polygons, the per-vertex weighted binary cross-entropy is
\begin{equation}
\mathcal{L}_{\mathrm{BCE}}(\phi) \,=\, -\frac{1}{\sum_{P \in \mathcal{B}} |V(P)|} \sum_{P \in \mathcal{B}} \sum_{i=1}^{|V(P)|} \big[\, w_{+} y_i \log \hat{p}_i + (1 - y_i) \log(1 - \hat{p}_i) \,\big].
\label{eq:setpred-loss}
\end{equation}
Whether closing that tail is the encoder's doing rather than the probe's own capacity is settled not here but by the controls of Section~\ref{sec:res-headline-ablation}: the no-encoder ablation and a zero-capacity linear probe.

\paragraph{Inference and the threshold knob} At inference we run the policy encoder once on the coordinates, greedy-decode the seed $\pi_{\mathrm{seed}}$, form the features $z_i$ of Eq.~\eqref{eq:setpred-feat}, run one \SetPredictor{} forward pass to obtain $(\hat{p}_1, \dots, \hat{p}_n)$, and threshold to return $S(t)$ (Eq.~\eqref{eq:setpred-decode}). The policy seed, the local-search editor, and the discretized-visibility scoring are all \emph{training-time} scaffolding used only to build the labels $S^\star_{\mathrm{LS}}$; at inference the method is this single geo-free forward pass --- no seed-refinement loop, no local search, no visibility object --- so LS is part of how the probe is \emph{trained}. The threshold $t$ is the single knob trading coverage for guard count, swept over $t \in \{0.20, 0.25, 0.30\}$ in Section~\ref{sec:res-dist}. Re-running the probe with its own output as the updated seed indicator changes nothing --- it is empirically a fixed point after the first pass (Section~\ref{sec:res-iteration}) --- so we use a single pass throughout.

\section{Experiments}
\label{sec:experiments}

\subsection{Dataset and partition}
\label{sec:exp-dataset}

We draw polygons primarily from the \emph{Art Gallery Problem with Vertex Guards} (AGPVG) instance library of Couto, de Rezende, and de Souza~\cite{art-gallery-instances-page,Couto2011ExactAGP}, which collects several polygon families --- random orthogonal (including large-area \texttt{fat} and small-area \texttt{min} variants), random simple (\texttt{randsimple}), random von Koch (\texttt{randvon}), complete von Koch (\texttt{vonkoch}), and boundary-simple (\texttt{simple}) --- over vertex counts $n$ ranging from $8$ to $2{,}500$. The library distributes vertex-guard optimal solutions for its instances, which we use directly as $\OPT$.

We supplement the library with $313$ additional random simple polygons (\texttt{random} class, $n \in \{20, 40, \dots, 600\}$, $30$ instances per size) to increase training density for random simple polygons in the in-distribution range; these share the same rational-coordinate format, and their vertex-guard optima were computed by solving the vertex-guard set-cover integer program. The splits we work with are listed in Table~\ref{tab:dataset_partition}.\footnote{The $70/30$ \texttt{dev}$/$\texttt{test} partition uses a seeded random shuffle (seed $1234$) rather than an alphabetical-name split due to instance naming}

\begin{table}[t]
  \centering
  \small
\begin{tabular}{lrrr}
  \toprule
  Split & \# polygons & $n$ range & Use \\
  \midrule
  train     & 8{,}867 & 8--198       & Pointer + SetPredictor training \\
  dev       & 840     & 16--198      & SetPredictor checkpoint selection \\
  test      & 362     & 8--192       & Held-out evaluation (in-distribution) \\
  ood       & 2{,}081 & 8--1{,}000   & Out-of-distribution evaluation \\
  ood-large & 285     & 600--2{,}250 & Extreme-OOD evaluation \\
  \bottomrule
\end{tabular}

  \caption{Dataset partition. The \texttt{dev} and \texttt{test} splits share the same source library files, partitioned $70/30$ by seeded random shuffle (seed $1234$): \texttt{dev} is used for checkpoint selection and \texttt{test} is the in-distribution held-out evaluation. The \texttt{ood} split is a larger held-out set spanning the training size range and extending well beyond it, to $n = 1000$ (about five times the training maximum); its genuinely larger-than-training polygons ($n > 198$) are analyzed separately in Section~\ref{sec:res-ood}. The \texttt{ood-large} split is used for an extreme-OOD evaluation (Table~\ref{tab:large}).}
  \label{tab:dataset_partition}
\end{table}

\subsection{Training setup}
\label{sec:exp-training}

The PO/BT policy is the existing \texttt{lstm\_bt} checkpoint: trained once from a single seed ($1234$) for $200$ epochs, with $R = 8$ rollouts per polygon, Bradley--Terry temperature $\alpha = 0.05$, and reward weights $\lambda = 1.0$ and $\rho = 3.0$ at the gate $\tau = 0.99$ (Eq.~\eqref{eq:reward}); all PO coverage rewards are computed from discrete-visibility sampling ($M = 500$), after which the encoder is frozen.

\medskip

\emph{Inputs:} a polygon's vertices are consumed by the LSTM in the source-library file order --- no canonical start vertex and no orientation (clockwise/counter-clockwise) are imposed --- and the coordinates are min--max normalized per polygon to $[0,1]^2$ before the encoder. The normalization makes the policy invariant to translation and to axis-aligned rescaling by construction; it is \emph{not} invariant to rotation or to a cyclic re-indexing of the vertices, whose empirical effect we quantify in Section~\ref{sec:limitations}.

\medskip 

\emph{Targets:} for each training polygon we obtain $S^\star_{\mathrm{LS}}(P)$ by running local search on the policy's seed under a coverage gate $\tau = 0.99$ against a discrete-visibility reference (Section~\ref{sec:method-pointer}); exact coverage-area computation is reserved for evaluation. These targets are \emph{not} produced by a policy-independent classical solver: they come from a best-improvement editor (add/remove/swap) initialized at the policy's own greedy-decoded seed and scored on discretized visibility. We refer to this policy-seeded refinement as ``LS'' throughout; it is the same edit family whose \emph{learned} counterpart we analyze in Section~\ref{sec:res-iteration}, here run to its best-improvement optimum as an oracle teacher rather than learned. 

\medskip 

\emph{Sizes:} the model dimensions of Section~\ref{sec:method-arch} take the following values:
\begin{center}\small
\begin{tabular}{ll}
  \toprule
  working width $d$                & $128$ \\
  self-attention blocks $L$        & $3$ \\
  attention heads $H$              & $8$ \\
  FFN expansion factor $f$         & $2$ \\
  \midrule
  policy parameters                & ${\approx}\,364\text{k}$ \\
  \SetPredictor{} parameters & $464{,}001$ \\
  \bottomrule
\end{tabular}
\end{center}

\medskip

\emph{Optimization:} weighted BCE (Eq.~\eqref{eq:setpred-loss}) with positive-class weight $w_{+} \approx 5.66$ (the empirical non-guard-to-guard ratio on \texttt{train}), $60$ epochs, batch size $32$, learning rate $3 \times 10^{-4}$, AdamW~\cite{Loshchilov2019AdamW}. Both the headline probe and the no-encoder ablation are four independent runs with seeds $\{1234, 11, 22, 33\}$ (reported as mean $\pm$ std); all figures display seed $1234$. Checkpoints are selected on \texttt{dev}; \texttt{test}, \texttt{ood}, and \texttt{ood-large} are touched only after training and threshold selection are complete.

\subsection{Evaluation protocol and metrics}
\label{sec:exp-eval}

For each polygon we greedy-decode the policy's $\EOS$-truncated seed and evaluate it with CGAL exact polygon visibility~\cite{cgal:eb-25b}; separately, we run one \SetPredictor{} pass, threshold at $t$ to obtain $S(t)$, and evaluate $S(t)$ with CGAL. Let $N$ be the partition size. We report mean coverage $\frac{1}{N}\sum \Covf{S}$, the normalized guard count $|S|/n$, and the approximation ratio $|S|/\OPT$ against the vertex-guard optimum. Because the mean hides the tail, we also report the per-polygon counts $\#\{\Covf{S} \geq c\}$ for $c \in \{0.95, 0.99, 0.999, 1\}$ and the worst case $\min_P \Covf{S}$; we call $\Covf{S} \geq \cfeas$ the \emph{feasibility} threshold and report the feasibility rate $\#\{\Covf{S} \geq \cfeas\}/N$ with Wilson $95\%$ confidence intervals. Where we compare the policy's and the probe's feasibility rates on the same polygons, we test the paired per-polygon outcomes with an exact McNemar test. The reporting threshold $c$ is distinct from the training-time gate $\tau = 0.99$ of Eq.~\eqref{eq:reward}. As noted in Section~\ref{sec:exp-dataset}, $\OPT$ values are taken from the vertex-guard optimal solutions distributed with the AGPVG library (for library instances) or computed via ILP (for the \texttt{random} class). For $79$ of the \texttt{ood-large} polygons (all with $n \geq 800$) no optimum is available --- most are library instances whose distributed solutions do not cover them (they are open at the source), and for the remainder our ILP did not terminate within budget. Restricting $|S|/\OPT$ to the $206$ solved instances would bias the ratio toward the easier $600$--$700$-vertex polygons, so on \texttt{ood-large} we report coverage and the normalized guard count $|S|/n$ over all $285$ polygons but no approximation ratio.

\paragraph{Computational cost} All experiments were run on a single workstation with an AMD Ryzen 9 3900X CPU, 64~GB RAM, and an NVIDIA GeForce RTX~2080~SUPER GPU (8~GB VRAM). Training the \SetPredictor{} probe is light: about $5$~s per epoch, $\approx 5$~minutes for a $60$-epoch run on the GPU above (the four-seed ensemble, $\approx 20$~minutes); the frozen PO/BT policy is reused rather than retrained. The geometric preprocessing --- the exact per-vertex CGAL visibility polygons and the $M=500$ discretized-visibility matrix --- is computed once per polygon and cached for the $\sim$$9.9$k library polygons, so it is not repeated across rollouts, edits, or seeds. At inference the method is a single geo-free forward pass: the full policy-encode-plus-probe pipeline runs in $\approx 25$~ms at $n=1000$ and $\approx 57$~ms at $n=2000$ on a GPU ($\approx 89$ and $\approx 264$~ms on CPU), the probe itself contributing only $\approx 10$~ms at $n=2000$. Exact CGAL coverage is used solely for evaluation and is the dominant offline cost, scaling with the guard-set size.

\section{Results}
\label{sec:results}

The results follow the hypothesis: the policy places guards but leaves a tail (Section~\ref{sec:res-headline}), reading the frozen encoder closes it (Section~\ref{sec:res-headline-ablation}), the effect generalizes out of distribution (Section~\ref{sec:res-ood}), and a single pass suffices because the probe is a fixed point under iteration (Section~\ref{sec:res-iteration}).

\subsection{The policy places guards}
\label{sec:res-headline}

\begin{table}[t]
  \centering
  \small
  \resizebox{\textwidth}{!}{
\begin{tabular}{lrrrr}
  \toprule
  Method & Mean cov & $\#\{\mathrm{Cov}\ge 0.95\}/N$ & Mean $|S|/n$ & Mean $|S|/\OPT$ \\
  \midrule
  \multicolumn{5}{l}{\emph{Classical baseline (non-learned)}} \\
  \quad Greedy & 0.9994 & 362/362 & 0.1524 & 1.0092 \\
  \midrule
  \multicolumn{5}{l}{\emph{Learned policy / probe}} \\
  \quad Pretrained pointer (seed) & 0.9689 & 293/362\,(0.766,0.847) & 0.1664 & 1.0885 \\
  \quad SetPredictor full ($t=0.20$) & \textbf{0.9985}\,$\pm$\,0.0010 & $361.5 \pm 0.9$/362 & 0.3888\,$\pm$\,0.0603 & 2.5560\,$\pm$\,0.4027 \\
  \quad SetPredictor, no-encoder ($t=0.20$) & 0.9972\,$\pm$\,0.0013 & $361.8 \pm 0.4$/362 & 0.4781\,$\pm$\,0.0504 & 3.1093\,$\pm$\,0.3350 \\
  \midrule
  \multicolumn{5}{l}{\emph{Oracle editor (\SetPredictor{} target, policy-seeded)}} \\
  \quad LS on policy seed & 0.9948 & 362/362 & 0.1369 & 0.8854$^{\dagger}$ \\
  \bottomrule
\end{tabular}
}
  \caption{Held-out evaluation on \texttt{test} ($362$ polygons). Classical greedy is the only non-learned anchor. The \emph{seed} row is the PO/BT policy decoded greedily; \emph{LS on policy seed} is its local-search refinement --- the \SetPredictor{}'s supervised target, not a classical baseline (Section~\ref{sec:exp-training}). \emph{\SetPredictor{} full} and \emph{no-encoder} are the headline probe and its zeroed-embedding ablation, each mean $\pm$ std over four seeds $\{1234, 11, 22, 33\}$ at $t = 0.20$ (Section~\ref{sec:res-headline-ablation}). Both probes saturate the $0.95$ gate ($\ge 361/362$); the encoder's effect surfaces at the stricter gates of Table~\ref{tab:dist_shift} and in the $\approx\!20\%$ higher guard cost the no-encoder probe pays ($|S|/\OPT$ $3.11$ versus $2.56$). Wilson $95\%$ CI shown only where the proportion is informative.\\
  $^{\dagger}$$|S|/\OPT < 1$ for the oracle editor because it halts at a disc-vis coverage gate ($\tau = 0.99$) and slightly undercovers (mean exact coverage $0.995$), so it uses fewer guards than the \emph{full-coverage} optimum. This is an artifact of the training reward, not a loose $\OPT$; the $|S|/n$ column is free of this caveat.}
  \label{tab:headline}
\end{table}

The geo-free policy on its own (Table~\ref{tab:headline}, \emph{seed} row) places guards at near-classical cardinality --- its mean $|S|/n$ ($0.166$) is close to classical greedy ($0.152$), at $|S|/\OPT \approx 1.09$ --- and clears the feasibility line on $293$ of the $362$ \texttt{test} polygons ($81\%$, Wilson $95\%$ CI: $0.766$, $0.847$). The LS oracle (bottom row) achieves $362/362$ feasibility at lower cardinality ($|S|/n = 0.137$, $|S|/\OPT = 0.885$), but is not geo-free: it reads discretized visibility at every add/remove/swap step and serves here only as the probe's training target, not as a geo-free competitor. The results show that vertex-guard placement is learnable, to a degree, by a method that never reads an explicit visibility object at inference. The remaining $69$ polygons ($19\%$) fall below the feasibility line under the policy alone; this tail is the paper's central motivating observation (the distribution is Section~\ref{sec:res-dist}), and the method gives no feasibility guarantee on its own. Figure~\ref{fig:worked_example} shows the seed, the seed plus probe at $t = 0.20$, and the optimum side by side for an in-distribution and an out-of-distribution polygon.

\figincl{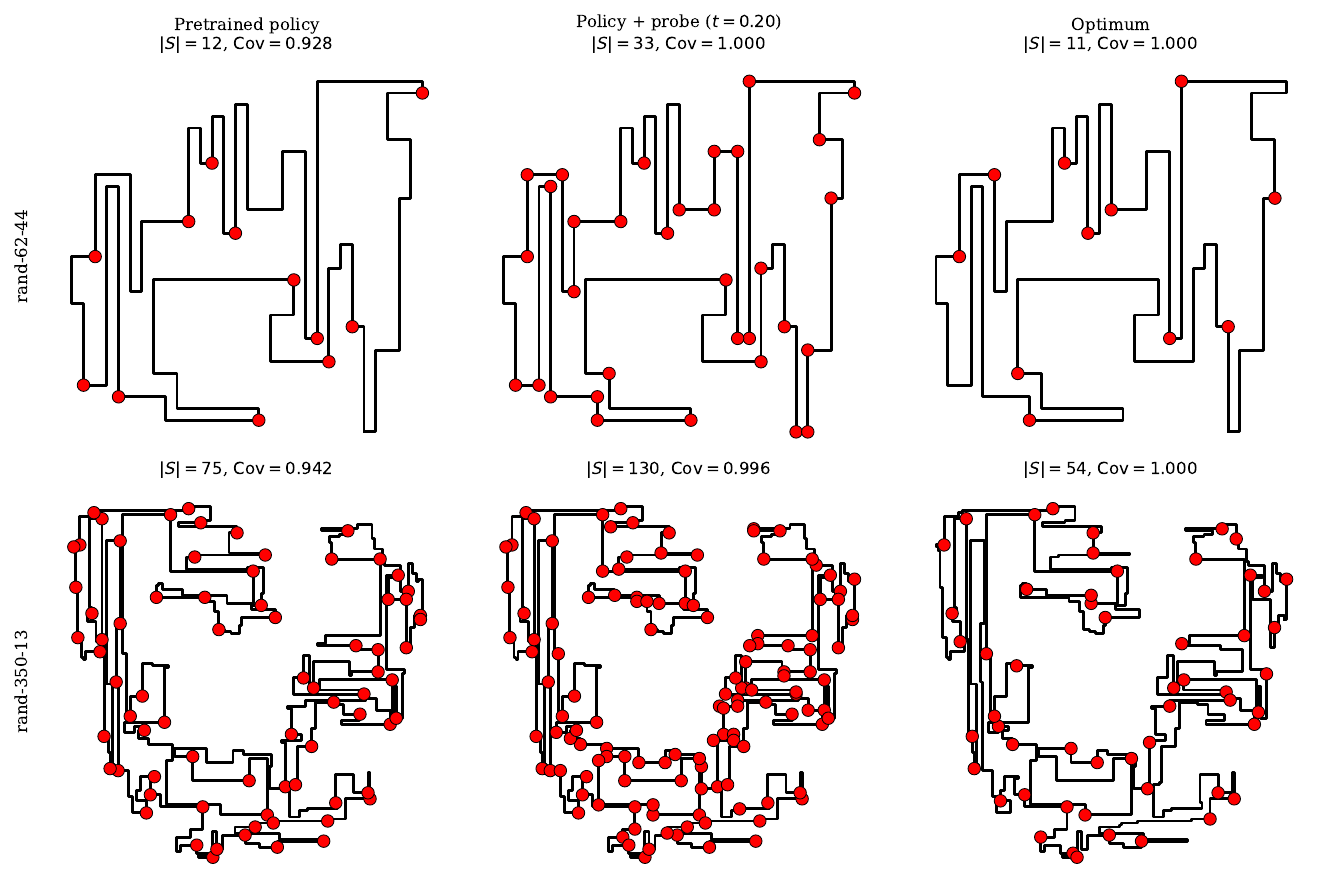}{Two polygons drawn with the policy's seed (left column), the policy plus the \SetPredictor{} probe at $t = 0.20$ (middle column), and the optimum (right column). Vertices selected as guards are marked. Row 1 is an in-distribution polygon where the policy already covers most of the region. Row 2 is an out-of-distribution polygon ($n = 350$, about $1.8\times$ the largest training polygon) where the policy's seed falls below the feasibility line ($\Covf{\cdot} = 0.942$) and the probe lifts it back above ($\Covf{\cdot} = 0.996$) at a guard cost typical for this split ($|S|/\OPT \approx 2.4$; cf.\ Table~\ref{tab:ood}).}{fig:worked_example}

\subsection{Reading the encoder closes the tail}
\label{sec:res-headline-ablation}
\label{sec:res-dist}

\begin{table}[t]
  \centering
  \small
  \resizebox{\textwidth}{!}{
\begin{tabular}{lrrrrr}
  \toprule
  Method & $\#\{\mathrm{Cov}=1\}$ & $\#\{\mathrm{Cov}\ge 0.999\}$ & $\#\{\mathrm{Cov}\ge 0.99\}$ & $\#\{\mathrm{Cov}\ge 0.95\}$ & min $\mathrm{Cov}$ \\
  \midrule
  Pretrained pointer (seed) & 3 & 15 & 97 & 293/362 & 0.830 \\
  SetPredictor $t=0.20$ (full) & 240 & 310 & 358 & 362/362 & 0.977 \\
  SetPredictor $t=0.20$, no-encoder & 120 & 163 & 306 & 362/362 & 0.952 \\
  SetPredictor $t=0.25$ (full) & 194 & 280 & 356 & 362/362 & 0.962 \\
  SetPredictor $t=0.30$ (full) & 151 & 219 & 350 & 362/362 & 0.952 \\
  \bottomrule
\end{tabular}
}
  \caption{Per-polygon coverage distribution on \texttt{test} ($362$ polygons). Counts are polygons reaching each coverage threshold; \emph{min} $\Covf{S}$ is the worst polygon. The policy has a long left tail; the \SetPredictor{} shifts the whole distribution right. This table exposes the encoder's contribution: where the $0.95$ gate saturates, the stricter $0.99$/$0.999$ gates and $\min\Covf{S}$ separate the full probe from the no-encoder ablation --- at $t=0.20$, $4$ polygons below $0.99$ versus $56$, and a worst polygon lifted from $0.952$ to $0.977$. All rows are the displayed seed ($1234$); Table~\ref{tab:headline} gives the four-seed mean$\pm$std.}
  \label{tab:dist_shift}
\end{table}

The policy's mean coverage hides a bimodal distribution (Table~\ref{tab:dist_shift}): it reaches full coverage on only a handful of polygons and trails a long left tail down to a worst polygon at $0.830$, while the \SetPredictor{} at $t = 0.20$ eliminates that tail and lifts the bulk above $0.99$. On the displayed seed the probe clears all $69$ of the policy's sub-feasibility polygons and introduces no new failures ($69 \to 0$ below $\cfeas$; paired exact McNemar test, $p \approx 3 \times 10^{-21}$); the same paired test is applied at OOD scale in Section~\ref{sec:res-ood}. The shift holds under the strict full-coverage requirement that defines the problem ($\Covf{S} = 1$): the probe covers $240$ of the $362$ polygons exactly against the policy's $3$ (Table~\ref{tab:dist_shift}), so the gain is not an artifact of reading feasibility at the $0.95$ gate.

The next question we ask is whether closing that tail is the encoder's doing or coordinates alone suffice. We isolate the encoder's contribution by retraining the probe with the frozen embedding $h_i$ masked to zero on every forward pass --- architecture, parameter count, optimizer, and training schedule identical (Section~\ref{sec:exp-training}). The $0.95$ feasibility gate does not reveal the encoder's contribution, as both probes saturate there ($362/362$, Table~\ref{tab:headline}) and the headline mean coverage is likewise nearly identical. The encoder's contribution lives in the tail. At the stricter $0.99$ gate of Table~\ref{tab:dist_shift}, on the displayed seed the full probe leaves $4$ polygons below against $56$ for the no-encoder ablation, a fourteen-fold difference, and lifts the worst polygon from $0.952$ to $0.977$ (across the four seeds the mean count below $0.99$ is $\approx 11$ for the full probe and $\approx 29$ for the no-encoder, of which the displayed seed's $56$ is the worst --- Table~\ref{tab:ablation}). The gap becomes qualitative out of distribution (Section~\ref{sec:res-ood}). This is the direct measurement behind C1 --- the frozen encoder carries geometric structure that raw coordinates do not reproduce, even with the decoder's full parameter budget.

The no-encoder ablation removes the encoder but keeps the policy's seed indicator among the probe's inputs, so the recovered feasibility could in principle be credited to the decoder's guard set rather than to the encoder. We therefore complete the $2\times2$ design, masking the seed indicator as well (Table~\ref{tab:ablation}, four-seed mean $\pm$ std at $t = 0.20$). The two axes must be read together, because at a fixed threshold the four conditions trade coverage against guard count differently. Masking the seed alone barely moves the probe: the encoder-only probe nearly matches the full probe both in the tail ($17.2 \pm 2.4$ versus $11.0 \pm 8.7$ polygons below $0.99$) and in cost ($|S|/\OPT$ $2.42$ versus $2.56$). Masking the encoder alone is markedly worse on the tail ($29.0 \pm 18.0$ below $0.99$ at $|S|/\OPT$ $3.11$). The coords-only probe (neither input) appears to have the smallest tail of all, but this is misleading: its threshold barely functions --- it selects $0.72$, $0.70$, and $0.67$ of all vertices at $t = 0.20, 0.25, 0.30$ --- so it holds coverage by \emph{flooding} guards rather than selecting them, never entering the guard-count regime the other probes operate in ($|S|/\OPT = 4.64$, against $2.4$--$2.6$). Reading the axes together, the seed contributes little, whereas the encoder is what makes economical feasibility reachable at all: it is the representation, not the decoder's seed, that lets the probe place few guards and still cover.

\begin{table}[t]
  \centering
  \small
\begin{tabular}{lrrrr}
  \toprule
  Probe inputs & $\#\{\mathrm{Cov}<0.99\}$ & $\min \mathrm{Cov}$ & $|S|/n$ & $|S|/\OPT$ \\
  \midrule
  \SetPredictor{} full & $11.0 \pm 8.7$ & 0.951\,$\pm$\,0.029 & 0.39\,$\pm$\,0.06 & 2.56\,$\pm$\,0.40 \\
  \quad no-seed (encoder only) & $17.2 \pm 2.4$ & 0.918\,$\pm$\,0.029 & 0.37\,$\pm$\,0.01 & 2.42\,$\pm$\,0.07 \\
  \quad no-encoder (seed only) & $29.0 \pm 18.0$ & 0.947\,$\pm$\,0.022 & 0.48\,$\pm$\,0.05 & 3.11\,$\pm$\,0.34 \\
  \quad coords-only (neither) & $0.8 \pm 0.8$ & 0.962\,$\pm$\,0.044 & 0.72\,$\pm$\,0.01 & 4.64\,$\pm$\,0.06 \\
  \bottomrule
\end{tabular}

  \caption{Encoder--seed ablation on \texttt{test} ($362$ polygons), all probes at $t = 0.20$, four-seed mean $\pm$ std over $\{1234, 11, 22, 33\}$. The four rows are the $2\times2$ on/off combinations of the two probe inputs --- the frozen encoder embedding and the policy's seed indicator; an absent input is masked to zeros, leaving architecture and parameter count unchanged. The two axes must be read together --- the coverage tail ($\#\{\mathrm{Cov}<0.99\}$, $\min\mathrm{Cov}$) against guard cost ($|S|/n$, $|S|/\OPT$) --- because the same threshold $t = 0.20$ does not produce the same operating regime across rows: a probe that selects nearly all vertices will show a short tail simply by over-guarding, not by genuinely learning to place guards. The encoder-only probe nearly matches the full probe on both; the seed-only probe (the no-encoder row of Table~\ref{tab:dist_shift}) widens the tail at higher cost. The coords-only probe is \emph{degenerate}: its threshold barely functions --- it selects $0.72$, $0.70$, $0.67$ of all vertices at $t = 0.20, 0.25, 0.30$ --- so it holds coverage by \emph{flooding} guards rather than selecting them, and its small $\#\{\mathrm{Cov}<0.99\}$ is an artifact of over-guarding ($|S|/\OPT = 4.64$), not selection quality. Only the encoder-bearing probes reach the economical regime ($|S|/\OPT \approx 2.4$--$2.6$); without the encoder the probe is stuck at $3.1\times$ (seed only) or $4.6\times$ (neither). The encoder is what makes economical feasibility \emph{reachable}.}
  \label{tab:ablation}
\end{table}

Since the \SetPredictor{} is itself a small Transformer, one might worry that \emph{it} computes the geometry and the encoder merely passes coordinates through. To rule this out we strip the probe of all capacity: a plain linear classifier (without any hidden layers or attention) reading only the frozen per-vertex features, asked to separate guard from non-guard vertices. It already reaches $\mathrm{ROC\text{-}AUC} \approx 0.84$, a measure of how reliably the rule ranks a guard above a non-guard ($0.5$ is chance, $1.0$ perfect).\footnote{$L_2$-regularized logistic regression under $5$-fold polygon-grouped cross-validation over $200$ polygons / $12{,}424$ vertices: $\mathrm{ROC\text{-}AUC} = 0.843 \pm 0.009$ and $\mathrm{PR\text{-}AUC} = 0.582 \pm 0.027$ (mean $\pm$ std across folds) against a $0.16$ positive-class prior.} A linear readout cannot be credited with computing geometry itself~\cite{AlainBengio2017Probes,HewittLiang2019Probes}, so the guard-relevant structure must already be linearly accessible in the frozen embedding --- the encoder, not the probe, carries it.  Scored by such a linear rule, guard vertices pile up at higher scores than non-guards, with only modest overlap, as visible in Figure~\ref{fig:embedding}.

\figincl[0.62\linewidth]{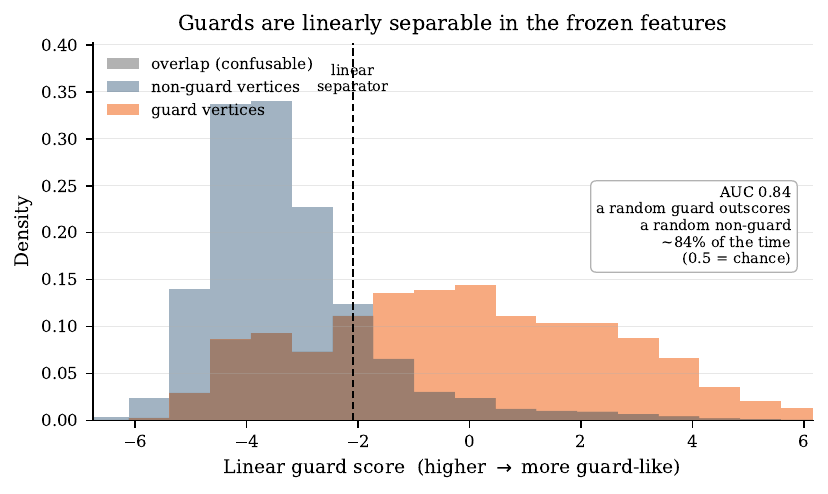}{Guard vertices are linearly separable in the frozen encoder features. A one-dimensional linear rule (LDA) over the same $200$-polygon / $12{,}424$-vertex probe set scores each vertex; the histograms show that score for guard and non-guard vertices separately. The rule is fit and scored \emph{out-of-fold} (GroupKFold over polygons --- no vertex is scored by a rule that saw its own polygon), so the separation reflects embedding structure rather than memorization. Guards score higher; the shaded band is the overlap the rule confuses. The separation corresponds to $\mathrm{ROC\text{-}AUC} \approx 0.84$, matching the logistic probe in the text.}{fig:embedding}

Closing the tail has a price, with the threshold $t$ as the knob: the approximation ratio rises as $t$ falls (Table~\ref{tab:pareto}), reaching $\approx\!2.6\times$ optimum at the headline threshold against $\approx 1.1$ for the policy alone. Figure~\ref{fig:distributions} gives the full picture in and out of distribution: the coverage CDF sits to the right of the policy at every threshold, while the $|S|/n$ and $|S|/\OPT$ box plots show the matching cost. We report the trade-off rather than fix one operating point.

\begin{table}[t]
  \centering
  \small
\begin{tabular}{lrrrr}
  \toprule
  Threshold $t$ & Mean $\mathrm{Cov}$ & Mean $|S|/n$ & Mean $|S|/\OPT$ & $\#\{\mathrm{Cov}\ge 0.95\}/N$ \\
  \midrule
  0.20 & 0.9993 & 0.4376 & 2.8972 & 362/362 \\
  0.25 & 0.9988 & 0.3802 & 2.5082 & 362/362 \\
  0.30 & 0.9979 & 0.3353 & 2.2041 & 362/362 \\
  \bottomrule
\end{tabular}

  \caption{Cost of feasibility on \texttt{test} ($362$ polygons), displayed seed ($1234$). The four-seed mean $|S|/\OPT$ is $2.02 \pm 0.21$ at $t = 0.30$ and $2.56 \pm 0.40$ at $t = 0.20$.}
  \label{tab:pareto}
\end{table}

\figincl{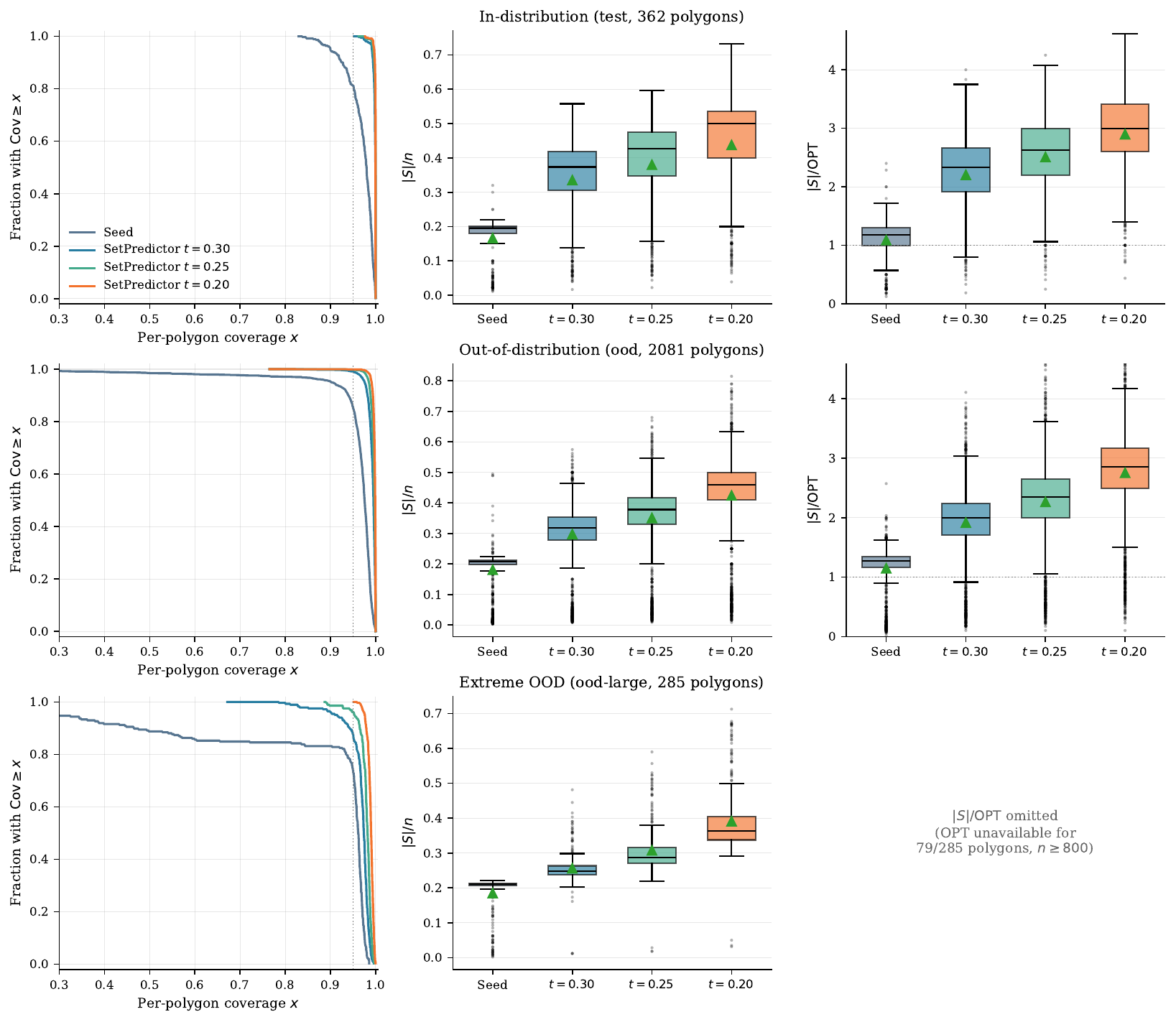}{Distributional view across all three regimes: \texttt{test} (top, $362$ polygons), \texttt{ood} (middle, $2081$), and \texttt{ood-large} (bottom, $285$, $n$ up to $2250$). \emph{Left:} per-polygon coverage as a complementary stair-step CDF; the dashed line marks the $0.95$ feasibility gate. \emph{Center:} box plots of $|S|/n$ for the policy seed and the three probe thresholds. \emph{Right:} the approximation ratio $|S|/\OPT$ for \texttt{test} and \texttt{ood} only; omitted for \texttt{ood-large}, where $\OPT$ is unavailable for the larger polygons (Section~\ref{sec:exp-eval}). Boxes show the displayed seed ($1234$); four-seed mean$\pm$std is in Tables~\ref{tab:headline} and~\ref{tab:ood}.}{fig:distributions}

\subsection{Out-of-distribution generalization}
\label{sec:res-ood}

\begin{table}[t]
  \centering
  \small
  \resizebox{\textwidth}{!}{
\begin{tabular}{lrrrr}
  \toprule
  Method & Mean cov & $\#\{\mathrm{Cov}\ge 0.95\}/N$ & Mean $|S|/n$ & Mean $|S|/\OPT$ \\
  \midrule
  Pretrained pointer (seed) & 0.9567 & 1778/2081\,(0.839,0.869) & 0.1807 & 1.1488 \\
  SetPredictor full ($t=0.20$) & \textbf{0.9946}\,$\pm$\,0.0020 & $2064.2 \pm 12.2$/2081 & 0.3617\,$\pm$\,0.0447 & 2.3341\,$\pm$\,0.2951 \\
  SetPredictor, no-encoder ($t=0.20$) & 0.9849\,$\pm$\,0.0055 & $2037.2 \pm 22.6$/2081 & 0.3667\,$\pm$\,0.0550 & 2.3583\,$\pm$\,0.3646 \\
  \bottomrule
\end{tabular}
}
  \caption{Out-of-distribution evaluation on \texttt{ood} ($2081$ polygons, $n$ up to $1000$). Rows as in Table~\ref{tab:headline}: the policy seed, the full probe, and the no-encoder ablation, both probes the four-seed mean $\pm$ std at $t = 0.20$.}
  \label{tab:ood}
\end{table}

The next test of the representation is the \texttt{ood} split, a larger held-out set that spans the training size range and extends well beyond it, reaching $n = 1000$ --- about five times the largest training polygon. Table~\ref{tab:ood} reports it. The policy alone drops to mean coverage $0.957$, with $303/2081$ polygons below the feasibility line ($85\%$ feasible). The probe at $t = 0.20$ (four-seed mean) cuts that to $16.8 \pm 12.2$ ($99\%$ feasible) at mean coverage $0.995$ --- roughly an order-of-magnitude reduction in failures. The improvement is significant for every probe seed individually (per-seed failure counts $3$, $8$, $22$, $34$; exact McNemar test on the paired per-polygon outcomes, $p < 1.2 \times 10^{-75}$ in all four cases), and the per-seed Wilson $95\%$ intervals for the probe's feasibility rate all lie above $0.977$. The no-encoder ablation leaves $44 \pm 23$ below feasibility (four-seed mean) at mean coverage $0.985$ --- about $2.6\times$ the full probe's failures (Table~\ref{tab:ood}), so out of distribution the encoder's contribution is qualitative, not the saturated tie it is in distribution. This is the evidence behind C2: a small classifier over the frozen embeddings recovers feasibility across this held-out set. The gain is not an artifact of the in-distribution-size majority of \texttt{ood}: on the genuinely larger-than-training polygons alone ($n > 198$, $885$ instances) the policy is feasible on $85\%$ and the probe on $98.7\%$ across four seeds ($100\%$ on the displayed seed). The reduction is robust across seeds, though the residual failure count is seed-dependent (the per-seed counts above), so at-scale generalization is partially seed-dependent. Figure~\ref{fig:distributions} (middle and bottom rows) shows the matching coverage and cost distributions for \texttt{ood} and \texttt{ood-large}.

\begin{table}[t]
  \centering
  \small
  \resizebox{\textwidth}{!}{
\begin{tabular}{lrrrr}
  \toprule
  Method & Mean $\mathrm{Cov}$ & $\min \mathrm{Cov}$ & $\#\{\mathrm{Cov}\ge 0.95\}/N$ & Mean $|S|/n$ \\
  \midrule
  Pretrained pointer (seed) & 0.8678 & 0.038 & 210/285 & 0.1849 \\
  SetPredictor full ($t=0.20$) & \textbf{0.9630}\,$\pm$\,0.0209 & 0.612\,$\pm$\,0.286 & $253.5 \pm 19.9$/285 & 0.2951\,$\pm$\,0.0579 \\
  SetPredictor, no-encoder ($t=0.20$) & 0.9295\,$\pm$\,0.0196 & 0.091\,$\pm$\,0.092 & $251.0 \pm 13.8$/285 & 0.3053\,$\pm$\,0.0477 \\
  \bottomrule
\end{tabular}
}
  \caption{Extreme out-of-distribution evaluation on \texttt{ood-large} ($285$ polygons, $n$ from $600$ to $2250$), coverage scored by exact CGAL at $t = 0.20$ over all $285$. The approximation ratio is omitted because $\OPT$ is unavailable for $79$ polygons ($n \geq 800$; Section~\ref{sec:exp-eval}). The full-probe and no-encoder rows are four-seed mean $\pm$ std; their $\min \mathrm{Cov}$ is the mean $\pm$ std of the four per-seed worst-polygon coverages.}
  \label{tab:large}
\end{table}

\paragraph{Extreme out-of-distribution} On the \texttt{ood-large} split ($n$ from $600$ to $2250$, up to about eleven times the largest training polygon), the worst-case coverage is the more informative metric, as the mean can look reasonable while the tail collapses (Table~\ref{tab:large}). The pretrained seed leaves its worst polygon at coverage $0.038$ with $75/285$ below feasibility, and the no-encoder ablation barely moves it: its worst-polygon coverage averages only $0.09 \pm 0.09$ across the four seeds (three of the four still pinned at $0.038$), so it trails the full probe on the mean ($0.930$ versus $0.963$) and collapses on the tail. The full probe lifts the worst polygon for every seed: across the four seeds the worst-polygon coverage at $t = 0.20$ averages $0.61 \pm 0.29$ (against the no-encoder's $0.09 \pm 0.09$), and on the displayed seed reaches $0.951$ with nothing left below feasibility. The seed dependence is genuine --- the four-seed feasibility count is $253.5 \pm 19.9$ of $285$, with the per-seed count below feasibility ranging from $0$ (the displayed seed) to $55$. At $t = 0.20$ the probe uses $|S|/n \approx 0.30$ guards on this split, the cost of restoring feasibility this far beyond the training range.

\subsection{Alternatives to the probe: decode search, editors, and the fixed point}
\label{sec:res-iteration}

\paragraph{Decode-time search does not close the tail} The simplest alternative to the probe is to decode the policy better rather than read its encoder. We test this directly (Table~\ref{tab:decode_search}): for each \texttt{test} polygon we draw $K = 32$ stochastic rollouts from the frozen policy and select among them geo-free, by length-normalized log-likelihood, with no visibility oracle. The result is statistically indistinguishable from greedy --- $294$ versus $293$ of $362$ feasible, $266$ versus $265$ below $0.99$, identical $|S|/\OPT \approx 1.09$ --- because the policy's maximum-likelihood decode already \emph{is} the greedy seed: likelihood-ranked search never prefers a more-covering set, and occasionally prefers a less-covering one (its worst polygon drops to $0.557$). Letting an \emph{oracle} pick the highest-coverage sample of the $32$ --- classical active search, no longer geo-free --- raises feasibility to $333/362$ but still leaves $199$ below $0.99$, far short of the probe ($362/362$ feasible, $\approx 4$ below $0.99$ on the displayed seed). The tail is thus not a decoding artifact that more search removes: closing it requires reading the encoder, and the probe pays for that in guards ($|S|/\OPT \approx 2.6$) rather than in search.

\begin{table}[t]
  \centering
  \small
  \resizebox{\textwidth}{!}{
\begin{tabular}{lrrrrrr}
  \toprule
  Decode of the frozen policy & $\#\{\mathrm{Cov}\ge 0.95\}/N$ & $\#\{\mathrm{Cov}<0.99\}$ & Mean $\mathrm{Cov}$ & $\min\mathrm{Cov}$ & $|S|/n$ & $|S|/\OPT$ \\
  \midrule
  greedy (seed) & 293/362 & 265 & 0.969 & 0.830 & 0.166 & 1.09 \\
  best-of-$32$, likelihood (geo-free) & 294/362 & 266 & 0.967 & 0.557 & 0.166 & 1.09 \\
  best-of-$32$, coverage oracle (\emph{not} geo-free) & 333/362 & 199 & 0.981 & 0.845 & 0.171 & 1.12 \\
  \bottomrule
\end{tabular}
}
  \caption{Decode-time search on \texttt{test} ($362$ polygons), all reading the same frozen PO/BT policy. \emph{greedy} is the seed of Table~\ref{tab:headline}. \emph{best-of-$32$, likelihood} draws $K=32$ stochastic rollouts and keeps the one with the highest length-normalized log-likelihood --- a geo-free decode search with no visibility oracle. \emph{best-of-$32$, coverage oracle} instead keeps the highest-coverage sample (tie-break: fewer guards); it consults exact coverage to choose and is therefore \emph{not} geo-free, reported only as an upper bound on what the policy's own samples contain. Greedy is included in each candidate pool, so neither selection can underperform it on its own ranking metric. Geo-free search matches greedy and does not close the tail; even oracle-assisted search falls well short of the \SetPredictor{} (Table~\ref{tab:dist_shift}).}
  \label{tab:decode_search}
\end{table}

Before settling on the single-shot probe we tried to improve the seed with a learned \emph{iterative} editor applying $\textsc{add}/\textsc{remove}/\textsc{stop}$ edits. Three architectures of increasing capacity --- including one with topology features and an auxiliary visibility-prediction loss, and one with DAgger refresh~\cite{RossGordonBagnell2011DAgger} --- failed to improve over the seed in any coverage-preserving regime. The best variant cut $|S|$ by about $24\%$ but lost coverage, reaching only a $0.27$ \emph{recovery rate} (the fraction of its edits that match or beat the local-search reference on both coverage and guard count) --- well below replacement quality. Three structural reasons recur: errors accumulate because the editor sees clean trajectories in training but its own predictions at inference; a single $\{\textsc{add},\textsc{remove},\textsc{stop}\}\times\{1,\dots,n\}$ head couples selection with termination, miscalibrating $\textsc{stop}$; and the train/inference state distributions diverge in ways DAgger refresh did not close.

The same teacher-forcing cascade also defeats a \emph{supervised} pointer network --- which we trained directly on AGPVG solutions before adopting the reinforcement recipe. Cloning a guard sequence is ill-defined for a second reason beyond the cascade: a polygon has many optimal and near-optimal guard sets (the constraint $\Covf{S}=1$ in Eq.~\eqref{eq:opt} fixes the cardinality, not the membership), so any single target sequence is an arbitrary choice among equivalent solutions, and an autoregressive loss penalizes the model for proposing a different but equally valid one. Forced onto sequences it would not itself generate, the recurrent decoder then drifts into hidden states unseen at training and collapses (mean coverage $\approx 0.50$ at roughly $3\times$ the optimum) --- the brittleness of supervised cloning anticipated in Section~\ref{sec:intro}. Preference optimization sidesteps both problems: it ranks the policy's \emph{own} rollouts by reward rather than matching a designated target, so non-unique optima are not in conflict. The single-shot probe likewise has none of these failure modes --- no rollout, no stop token, and a per-vertex membership label rather than an ordered sequence, trained and used in the same regime.

A clarification this invites: the probe's supervised targets $S^\star_{\mathrm{LS}}$ are themselves the output of this same edit family run to its best-improvement optimum --- an \emph{oracle} editor seeded by the policy. So the contrast is sharper than ``editors fail, the probe works'': the \emph{learned} editor fails to reproduce the oracle's edits autoregressively, whereas a single-shot probe trained on the oracle's endpoints recovers them in one pass. That the probe succeeds is therefore not, by itself, evidence about the encoder --- it could merely be distilling the oracle. What attributes the success to the \emph{representation} is the encoder--seed ablation of Section~\ref{sec:res-headline-ablation} (Table~\ref{tab:ablation}), run with the LS target held fixed: masking the seed indicator barely moves the probe, masking the frozen embedding markedly widens the tail, and removing both leaves a probe that stays feasible only by flooding guards ($|S|/\OPT \approx 4.6$); a zero-capacity linear probe over the same embedding already separates guards (ROC-AUC $0.843$). Were the probe merely distilling the oracle, which inputs it reads would not matter --- yet it is the encoder, not the decoder's seed, that makes the targets predictable from coordinates alone.

Re-running the probe with its own output as the updated seed indicator does not change the prediction in any meaningful way: across $K \in \{1, 2, 3, 5\}$ the three metrics barely move (Figure~\ref{fig:mechanism}). Almost all of the movement is a single settle between $K = 1$ and $K = 2$ at the highest threshold ($t = 0.8$) --- at most ${\approx}\,0.02$ in $|S|/\OPT$ ($\approx 0.011$ in coverage, $\approx 0.004$ in $|S|/n$); beyond $K = 2$ nothing changes by more than ${\approx}\,8.5 \times 10^{-3}$ (in $|S|/\OPT$; coverage and $|S|/n$ stay under $2.5 \times 10^{-3}$). This is the fixed-point property (C4): with the seed indicator already among its inputs, a single pass suffices.

\figincl{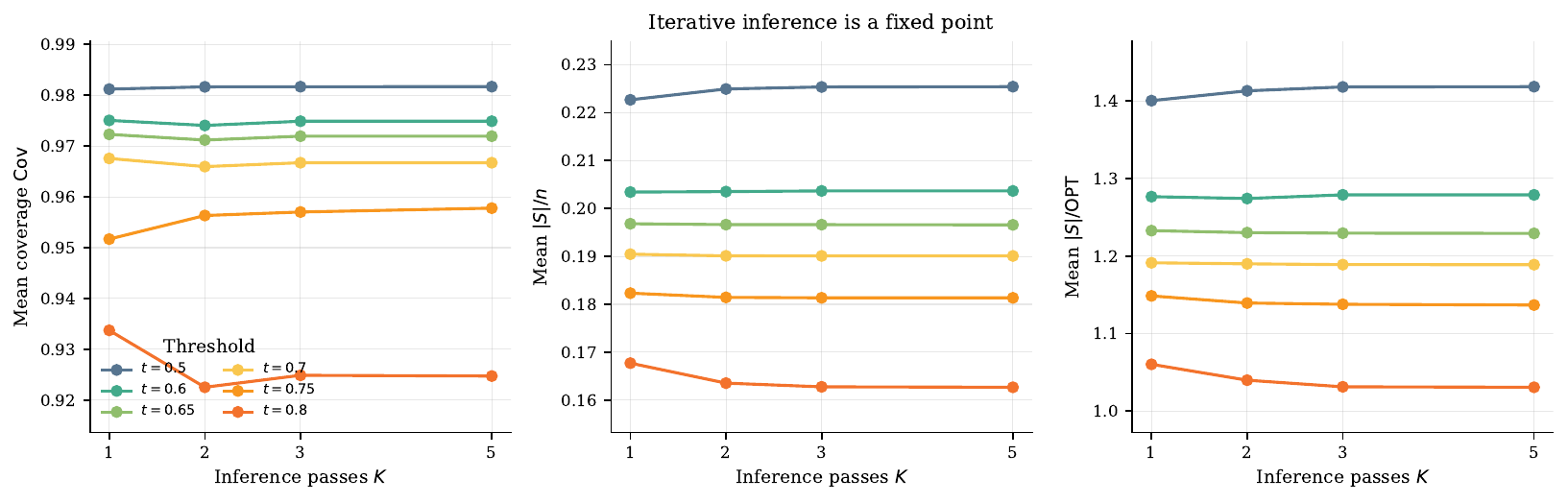}{Iterative inference is a fixed point across all three metrics, on the combined \texttt{dev}$+$\texttt{test} pool ($1224$ polygons). Each panel shows mean coverage (\emph{left}), $|S|/n$ (\emph{centre}), and $|S|/\OPT$ (\emph{right}) versus inference passes $K \in \{1,2,3,5\}$ for six thresholds $t \in \{0.5, 0.6, 0.65, 0.7, 0.75, 0.8\}$ --- a higher range than the headline thresholds, where the fixed-point property is most stringently tested near the decision boundary. The metrics barely move (bounds in Section~\ref{sec:res-iteration}); almost all of the change is a single settle between $K=1$ and $K=2$ at $t=0.8$.}{fig:mechanism}

\section{Discussion}
\label{sec:discussion}

The central measurement in this paper is what happens when we read the encoder rather than the decoder. Reading only the frozen embeddings, the vertex coordinates, and the policy's seed indicator --- no visibility matrix, no CGAL output, no per-vertex area feature --- the \SetPredictor{} closes the in-distribution feasibility tail on the displayed seed (Table~\ref{tab:headline}) and compresses the out-of-distribution one by roughly an order of magnitude across four probe seeds (Section~\ref{sec:res-ood}). Two readings are consistent with this. Either the encoder learned enough geometric structure that a small classifier over its output recovers feasibility on its own, or the decoder's $\EOS$ calibration was the bottleneck and the policy had already identified the right vertices but stopped too early. Both point to the same headline conclusion --- the reinforcement method internalized a representation that holds more of the relevant geometry than its decoder expressed --- and the encoder--seed ablation (Table~\ref{tab:ablation}) adjudicates between them in the encoder's favour: masking the decoder's expressed seed barely changes the probe, whereas masking the encoder widens the tail or forces wholesale over-guarding, so the carried signal is the encoder's geometry rather than the decoder's stopping behaviour. Three measurements pin this down (Section~\ref{sec:res-headline-ablation}). The no-encoder ablation --- masking the embedding to zeros with everything else held constant --- leaves a probe that cannot match the full probe's coverage at matched cardinality on either split, separating most clearly at the $0.99$ gate of Table~\ref{tab:dist_shift}. The complementary no-seed control shows the encoder alone nearly recovers the full probe, so the seed is not what carries the signal. And the linear probe (ROC-AUC $0.843 \pm 0.009$) shows the guard-relevant structure is linearly accessible in the frozen embedding, with Figure~\ref{fig:embedding} as the matching visualization.

Recovering feasibility has a price, and we report it rather than disguise it. The probe's guard sets grow to roughly two to three times the optimum as the threshold is lowered (Section~\ref{sec:res-dist}), and the $|S|/\OPT$ box plots of Figure~\ref{fig:distributions} show that the whole distribution widens, not just its mean. A deployment would pick one operating point on this curve; we leave it as a knob because the right point depends on how costly an extra guard is relative to an uncovered region.

Note that we are not claiming a better AGP solver. Classical greedy reaches mean coverage $0.999$ at $|S|/n = 0.152$ and $|S|/\OPT = 1.009$ on \texttt{test} (Table~\ref{tab:headline}); the probe at $t=0.20$ reaches mean coverage $\geq 0.998$ at $|S|/\OPT \approx 2.6$ across four seeds. A greedy-then-local-search pipeline --- the standard high-quality classical recipe --- would be stronger still on cardinality; the \emph{LS on policy seed} row of Table~\ref{tab:headline}, which refines a guard set with the same add/remove/swap editor to $362/362$ feasibility at $|S|/\OPT = 0.885$, already bounds what such refinement achieves on this data. The classical anchors win on cardinality at preserved coverage, and the policy and probe are not built to contest that. What we document is what is learnable by reinforcement when the model is denied geometric input at the moment of placing guards, and what part of that learning lives in the representation rather than the decoder. The contrast between the failed iterative editors (Section~\ref{sec:res-iteration}) and the single-shot probe is the cleanest expression of the point: removing the stop time and the rollout is exactly what lets the encoder's information surface.

\section{Limitations}
\label{sec:limitations} 

We acknowledge several limitations of the work, enumerated here for clarity; the first two are conceptual and the rest experimental.

\begin{enumerate}
  \item \textbf{Scope of the ``no geometric knowledge'' claim.} The geo-free constraint is on inference: the policy and the probe see no visibility object at test time. \emph{Training} does see visibility: the PO/BT reward is computed from discrete visibility coverage, and the probe's supervised targets are generated by local search using discrete visibility as a coverage gate. ``No explicit geometric knowledge'' must be read as ``no geometric oracle at the moment of placing guards.''
  \item \textbf{The probe is supervised, not reinforcement.} The \SetPredictor{} is trained by BCE against local-search-derived labels. It is not itself a reinforcement learner. We position it as a representation probe of the RL-trained encoder, not as an RL contribution.
  \item \textbf{Held-out partition size.} \texttt{test} has $362$ polygons. Wilson $95\%$ CIs on the coverage-feasibility proportion are reported, but the absolute count is modest. The OOD evaluation on \texttt{ood} ($2081$ polygons) provides a much larger second held-out evidence base.
  \item \textbf{Policy selection and seeds.} The PO/BT policy is the best of several reinforcement-pretraining runs, selected by training-set performance. Because selection used only the \texttt{train} split, \texttt{test} and \texttt{ood} play no role in it and the held-out evaluations are not optimistically biased by the choice; selecting on training reward rather than a held-out criterion is a weaker rule that, if anything, risks favoring an over-fit run, so we make no claim the selected policy is optimal. We do not average over policy seeds; seed-variance is characterized separately for the probe over four seeds $\{1234, 11, 22, 33\}$ (Table~\ref{tab:headline}, Section~\ref{sec:res-ood}).
  \item \textbf{LSTM, not Transformer, policy.} We trialed a Transformer encoder--decoder policy of roughly $6\times$ the parameters; it reached a comparable coverage--cost operating point (greedy coverage $\approx 0.97$ at $|S|/\OPT \approx 1.15$ versus the LSTM pointer's $\approx 1.21$ --- sparser but no better on the trade-off), one training run failed to converge, and we found no improvement justifying the added capacity. We read this as overparametrization relative to the scalar RL signal rather than evidence against attention models in general --- consistent with the fact that the probe, a small Transformer trained on dense per-vertex labels, does work; a careful attention-model study is future work.
  \item \textbf{Decode-time search and stronger decoders.} We tested decode-time search of the policy seed by best-of-$K$ stochastic sampling with length-normalized scoring (Section~\ref{sec:res-iteration}, Table~\ref{tab:decode_search}): geo-free likelihood-ranked search matched greedy and did not close the tail, so the residual tail is not a decoding artifact. We did not implement exhaustive beam search --- the decoder loop is not exposed for per-step beam expansion, and best-of-$K$ sampling is the stronger search in practice --- nor symmetry-exploiting RL such as POMO~\cite{Kwon2020POMO}; these act on the decoder and are orthogonal to the representational question of what the encoder has learned, so we leave them to future work.
  \item \textbf{Discrete visibility approximation in training.} Both the PO/BT coverage reward and the \SetPredictor{} training targets use discrete visibility sampling at $M = 500$ (Section~\ref{sec:method-pointer}). Exact coverage-area computation is used only at evaluation; the per-vertex visibility polygons are computed with CGAL in both training and evaluation. We did not sweep $M$: the sample average has standard error $O(1/\sqrt{M})$, so a larger $M$ --- or stratified sampling near reflex vertices, where the uncovered slivers are hardest to detect --- would reduce target noise, but quantifying that sensitivity is left to future work.
  \item \textbf{Reward-hyperparameter sensitivity.} The reward weights and PO settings ($\lambda = 1.0$, $\rho = 3.0$, $R = 8$ rollouts, $\alpha = 0.05$; Section~\ref{sec:exp-training}) were fixed to the values that produced the released checkpoint rather than swept. The design turns on the \emph{form} of the reward --- capping coverage at $\tau$ and linearly penalizing the deficit (Section~\ref{sec:method-pointer}) --- rather than on the precise scalars; a sensitivity study over them would require re-running PO/BT training and is left to future work.
  \item \textbf{Extreme OOD is strong but seed-dependent.} On the \texttt{ood-large} split (Table~\ref{tab:large}, Section~\ref{sec:res-ood}) the frozen-encoder probe lifts worst-case coverage on every seed while the no-encoder ablation cannot, but the per-seed feasibility count is variable, so we read it as evidence of the encoder signal rather than a feasibility guarantee at this scale; the probe also pays roughly double the local-search guard count, and the vertex-guard optimum is unavailable for the largest polygons there.
  \item \textbf{Training-curve reconstruction.} Figure~\ref{fig:po_training} is reconstructed from four saved checkpoints (epochs $110$, $114$, $160$, $200$), not from an online metric log; per-epoch coverage was not preserved during the original PO/BT run. The trace shows only late-training dynamics, which is sufficient to confirm that the gradient direction does not collapse but does not characterize early-training behavior.
  \item \textbf{Invariance and vertex ordering.} The recurrent encoder reads vertices in file order, so the pipeline is invariant to translation and axis-aligned scaling by construction (the input is min--max normalized; Section~\ref{sec:exp-training}) but not, in principle, to rotation, reflection, or vertex re-indexing. We measured the effect on \texttt{test} by re-running the frozen policy and probe ($t = 0.20$) on transformed polygons, re-decoding the seed each time: under rotation ($45^\circ$, $90^\circ$, $180^\circ$) and mirroring the feasibility rate is unchanged ($362/362$ at the $0.95$ gate, mean coverage $\geq 0.998$), so in aggregate the learned features tolerate these transforms; a cyclic re-indexing of the vertices is slightly more disruptive ($359/362$ feasible, with one small polygon falling to coverage $0.79$), consistent with the order-dependence a recurrent encoder introduces. A permutation- or rotation-equivariant encoder (for instance, attention with relative positional encoding) would remove this dependence by construction and is a natural next step.
\end{enumerate}

\section{Conclusion}
\label{sec:conclusion}

We asked what a neural policy internalizes about vertex-guard placement when trained from a coverage-aware reward over its own rollouts and denied any geometric oracle at inference, and we answered it by reading the encoder rather than the decoder: a single-shot probe over the frozen embeddings closes the in-distribution feasibility tail on the displayed seed and compresses the out-of-distribution tail by roughly an order of magnitude across four probe seeds, which we take as evidence that the reinforcement-trained encoder had captured more of the placement geometry than its decoder emitted as guards. The result is a measurement in a deliberately constrained setting, not a competitor to classical solvers on guard count --- the learner is not the best AGP heuristic, but it arrived at a usable representation of vertex-guard placement without ever being shown the geometry it had to reason about at inference. Two extensions follow naturally: a stronger encoder, such as a Transformer trained under the same PO/BT objective, may shorten the residual tail directly and leave less work for the probe; and a per-instance threshold, predicted geo-free from the same embeddings but trained against coverage-selected targets exactly as the probe is trained against the visibility-derived labels $S^\star_{\mathrm{LS}}$, would replace the global knob with an adaptive one --- keeping visibility a training-time target, never an inference-time input. More broadly, the encoder-probing decomposition is not specific to vertex guarding: it should transfer to other geometric set-cover problems whose feasibility oracle is costly at inference --- terrain guarding or range-limited watchtower placement, for instance --- where geo-free inference stays meaningful precisely because the visibility or coverage oracle one would otherwise consult is expensive to evaluate. Whether a frozen encoder carries the requisite structure in those settings is an open question the present method is designed to ask.

\section*{Data and Code Availability}

The AGPVG instance library is publicly available~\cite{art-gallery-instances-page}. The $313$ supplemental \texttt{random}-class polygons with their ILP-computed vertex-guard optima, the dataset splits, the trained checkpoints, and all training and evaluation code are available at \url{https://github.com/dseverdi/AGNet}.

\section*{Conflict of Interest}

The authors declare no competing interests.

\bibliographystyle{elsarticle-num}
\bibliography{paper}
\end{document}